# Fine-Tuning VGG Neural Network For Fine-grained State Recognition of Food Images

Kaoutar Ben Ahmed, Ahmad Babaeian Jelodar

*Abstract*—State recognition of food images can be considered as one of the promising applications of object recognition and fine-grained image classification in computer vision. In this paper, evidence is provided for the power of convolutional neural network (CNN) for food state recognition, even with a small data set. In this study, we fine-tuned a CNN initially trained on a large natural image recognition dataset (Imagenet ILSVRC) and transferred the learned feature representations to the food state recognition task. A small-scale dataset consisting of 5978 images of seven categories was constructed and annotated manually. Data augmentation was applied to increase the size of the data.

I. INTRODUCTION

Food recognition has recently started to gain attention and several works have been published [1][2][3][4][5]. The importance of automatic food recognition comes from its usefulness in dietary assessment tools. Instead of manual reporting of the patients' food intakes, automatic recognition of food can be more accurate and time saving.

Recent state of the art results have been obtained using deep learning in the fields of computer vision, object recognition, speech recognition, bioinformatics, and others. By training a neural network, we can drastically reduce the amount of time required for feature extraction. Besides, deep features can generalize very well to new scenarios compared to handcrafted features. Also, neural networks can extract features that annotators can miss. The most popular deep learning models used to train image datasets are convolutional neural networks (CNN). A CNN usually consists of a number of convolutional and subsampling layers followed by fully connected layers. In a Convolutional Neural Network, the input data x has a spatial structure: $x_l \in R^{H_l \times W_l \times C_l}$ where $H_l$ (height) and $W_l$ (width). The third dimension $C_l$ refers to the number of image channels (equals to 3 for Red-Green-Blue colored images). There exists a fourth dimension $N_l$ that corresponds to the batch size, which are subsets of data grouped in a single batch to perform batch processing. CNNs have shown to be very effective in learning features with a high level of abstraction when using deeper architectures and fine-tuned parameters. Since they were first introduced in [6], CNNs were improved and made deeper over the years. Recently, deep CNNs trained using large labeled datasets such as ImageNet [7] achieved the best results on the most challenging large image classification and detection datasets. There are numerous architectures in the field of Convolutional Neural Networks that have a recognized name we mention:

AlexNet. It was the first work that made Convolutional Networks popular in Computer Vision, developed by Alex Krizhevsky, Ilya Sutskever and Geoff Hinton. AlexNet was evaluated in the ImageNet ILSVRC challenge in 2012 and it had very good results (top 5 error of 16%).

VGGNet. The runner-up in ILSVRC 2014 developed by Karen Simonyan and Andrew Zisserman. Their pre-trained model is available in Caffe.

Practically, it is rare to attempt to train an entire Convolutional Network from scratch (especially with random initialization), due to the difficulty of finding a training dataset of large size. However, it is common to pre-train a CNN on a large-scale dataset, and then use the CNN for the target task (in our case food state recognition). The CNN can be used as fixed feature extractor or as an initialization. In order to use the CNN as fixed feature extractor we take a CNN pre-trained on a large dataset (e.g. ImageNet), remove the last fully connected layer, then use the rest of the CNN as a fixed feature extractor for the new dataset. There is another method to avoid training a CNN from scratch, which is to not only replace and retrain the classifier on top of the pre-trained CNN on the new dataset, but to also fine-tune the weights of the pre-trained network by continuing the back propagation. It is possible to fine-tune all the layers of the CNN, or it's possible to fix some of the earlier layers unchanged and only fine-tune some later layers of the network. This is motivated by the studies that noticed the generalization of the earlier features of a CNN and that they should be useful to different tasks, but later layers of the CNN become more specific to the details of the original dataset. The size of the new dataset (smaller or bigger), and its similarity to the original dataset are two important factors to be considered when deciding the adequate type of transfer learning. For instance, if the new dataset is small, it is not a good idea to fine-tune the CNN due to over-fitting concerns. If the new dataset is similar to original dataset, we expect higher-level features in the CNN to be relevant to this dataset. Thus, it is recommended to train a linear classifier on the CNN codes. Otherwise, if the new dataset is large then we might try to fine-tune through the full network. However if the new dataset is different from the original dataset and it has a small size, then it might work better to train a linear classifier from activations somewhere earlier in the network.

Kaoutar. B. Ahmed and Ahmad B. Jelodar are with the Computer Science and Engineering Department, University of South Florida, Tampa, FL 33620 USA. (E-mail: kbenahmed@mail.usf.edu, ajelodar@mail.usf.edu).

This study was aimed to recognize food state (sliced, grated, whole ... etc.) by applying a fine-tuned pre-trained CNN aided with data augmentation to a seven-class small-scale food image data. In robotics, cocking objects at different states require different instrument manipulations and grasping [8]. Therefore, object state recognition will definitely help robots to better perform manipulation-oriented grasp. Robots can be made to decipher a task goal, seek the correct objects at the desired states on which to operate, and generate a sequence of proper manipulation motions [9].

The remainder of this paper is organized as follows. Section II is a review of some of the related work. Section III presents the dataset used and the preprocessing performed. Section IV explains details of the methodology used in this study. Section V discusses the results obtained. Finally, Section VI consists of conclusions based on the experimental results.

## II. RELATED WORK

In their work [1], Yanai et al. used different variations of CNN for food recognition. They tried pre-training with the large-scale ImageNet data, fine-tuning and activation features extracted from the pre-trained model. The best performance achieved was 78.77%.

Liu et al. [10] created a food recognition system, which sends food images captured by the user's device to a server where image preprocessing and segmentation is done. The images are then fed into a convolutional neural network built using the inception modules first introduced by GoogLeNet [11]. In addition they designed a real-time food recognition system employing edge computing service paradigm. Their approach achieved 77% top1 accuracy.

Hou et al. [12] proposed a CNN-based algorithm for fruit recognition from image regions, which were extracted using selective search algorithm. Their model contains three convolutional layers, each of them is followed by pooling layers, and two fully connected layers. The dataset used consists of 5330 fruit images. 4000 images are used for training, 1330 for testing. The model had an accuracy of 99.77%.

Pan et al. [13] explored Multi-class classification of food ingredients. This model combines ResNet deep feature sets, Information Gain (IG) feature selection, and the SMO classifier. Their approach was evaluated on a multi-class dataset that includes 41 classes of food ingredients and 100 images for each class. The maximum accuracy achieved was 87.78%.

Rohrbach et al. [14] introduces a dataset, which distinguishes 65 fine-grained cooking activities such as cut and peel, or at an even finer scale namely cut slices and cut dice. The dataset includes high resolution image and video sequences. The study attempted to recognize the activities using one-vs-all SVMs and combined features namely trajectory, HOG, HOF and MBH. The highest accuracy obtained was 59.2%.

The last work review suggests that fine-grained recognition is clearly beyond the current state-of-the-art and further research is required to tackle this challenge. Additionally, to our knowledge no previous work has been done to recognize fine-grained state of food (i.e. sliced, diced, juice…etc.)

## III. DATASET

The dataset includes approximately 18 cooking objects (tomato, onion, bread, pepper, cheese, ...etc.) with 7 different states (whole, cut, sliced, chopped, grated, paste, juice). The dataset was manually annotated by human subjects. Each subject was assigned 10-20 short videos (~ 1 minute of 0.5-1 fps) of only a single object (for example: onion). The subject had to draw the bounding box for each object of the type (onion) in each frame of the video and annotate the bounding box with the state of that object (for example: sliced).

Figure 1. Examples of different food states included in the dataset

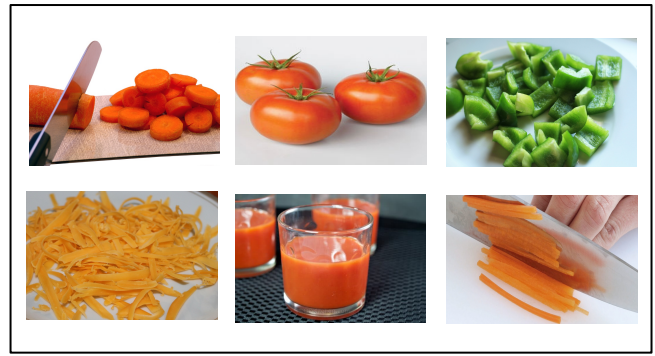

To ensure the quality of annotations, each subject was given a random set of videos and images from other subjects to check for annotation errors. Finally, the constructed dataset consists of a total of 5978 annotated images.

Since the pre-trained CNN we used requires an input image size of 224×224, we resized all images into 224 × 224, by using the imresize function in Matlab.

Given the limited annotated images available, data augmentation technique is used to increase the size of the available data. It is a set of label- preserving transformations to produce new instances without affecting the semantic of class label. Therefore, affine transformations including rotation and flipping were conducted to expand the training data. Oversampling was also used to balance the dataset since some classes had significantly less instances than others. A total of 10547 augmented images were produced. 90% were used for training and 10% for validation. For normalization purposes, the mean of the training set is subtracted from all images including the validation and testing sets. The test set consists of 861 unseen images from all seven classes.

## IV. METHODOLOGY

The pre-trained CNN we used in this study was the Fast (CNN-F) architecture presented in Chatfield's work [15]. As shown in Table I, CNN-F architecture has 5 convolutional layers followed by 3 fully-connected layers, resulting in a total of 8 learnable concatenated layers. The parameters in the table specifies the number of convolution filters and their receptive field size as "num x size x size", the convolution stride and spatial padding. "LRN" indicates if Local

Response Normalization (LRN) was applied and "pool" indicates the max-pooling down sampling factor.

The pre-trained model we used is VGG-F provided by MatconvNet [16], which is a MATLAB toolbox implementing Convolutional Neural Networks (CNNs) for computer vision applications. The deep model was pre trained on ILSVRC-2012 with the ImageNet dataset with momentum 0.9; weight decay $5 \cdot 10^{-4}$; initial learning rate $10^{-2}$, which is decreased by a factor of 10, when the validation error stops decreasing even though the training error keeps decreasing. Gaussian distribution with a zero mean and variance of $10^{-2}$ was used to initialize layers. The pre-training was performed on a single NVIDIA GTX Titan GPU and it run for 5 days. The activation function for all weight layers (except for fc8) is the Rectification Linear Unit (RELU) defined as $f(x)= \max(0, x)$ Where x is the input to a neuron. The required input is an RGB image with a size of 224×224 pixels.

TABLE I. CNN-F ARCHITECTURE.

| Layer | Parameters |
|---|---|
| conv1 | 64×11×11 stride 4, pad 0, LRN, x2 pool |
| conv2 | 256×5×5 stride 1, pad 2, LRN, x2 pool |
| conv3 | 256×3×3 stride 1, pad 1 |
| conv4 | 256×3×3 stride 1, pad 1 |
| conv5 | 256×3×3 stride 1, pad 1, x2 pool |
| fc6 | 4096 dropout |
| fc7 | 4096 dropout |
| fc8 | 1000 Softmax |

A common practice for fine-tuning is to replace the last fully connected layer of the pre-trained CNN with a new fully connected layer that has as many neurons as the number of classes in the target task. We change the size of the last output layer to the same number as the number of the food states. In the experiments, we set the number of food states to 7; therefore, the new fully connected layer has 7 neurons. We randomly initialize the weights of the new layer.

The CNN was fine tuned with stochastic gradient descent. We fixed the weights of the five convolutional layers of the pre-trained network and only fine-tuned succeeding layers. This is motivated by the studies that noticed the generalization of the earlier features of a CNN and that they should be universal and beneficial to different tasks, but later layers of the CNN become more specific to the details of the original dataset. Especially with small training datasets, it may be a good idea to partially tune the network instead of full tuning due to over-fitting concerns. So, we only tuned the fully connected layers. We also tried to avoid over-fitting by adding two layers of 50% dropout after fc6 and fc7 layers. Dropout randomly turns off network connections at training time to fight over-fitting. In order to increase the stability of the model, speed up learning and for regularization purposes, batch normalization was added after conv1 and conv2 layers.

Training a DNN requires to select a set of hyper parameters, among which the leaning rate (η) is the most critical one affecting the training performance. A predefined learning rate over the entire training epochs seems suboptimal, since for instance a high learning rate will cause the parameter vector to bounce around indefinitely. Here, a dynamically updated learning rate was used. The learning rate set to be less than the initial learning rate used to optimize the CNN for ILSVRC12. This ensures that the features learned from the larger dataset are not completely forgotten.

The batch size was set to 50, momentum to 0.5 and number of epochs to 400. Different optimizers were tested including Adam and Rmsprop. However, stochastic gradient descent was the most suited for this dataset. The finetuning of the networks ran on a GeForce GTX 960 GPU using Cuda implementation of MatconvNet functions. The computing GPU nodes were provided by CIRCE Compute Cluster in University of South Florida.

## V. RESULTS AND DISCUSSIONS

Figure 2 shows the evolution of the top1 loss of the tuned neural network across iterations. The graph provides the top1 error of both training and validation sets.

Figure 2. Top1 error graph of the training vs. validation sets.

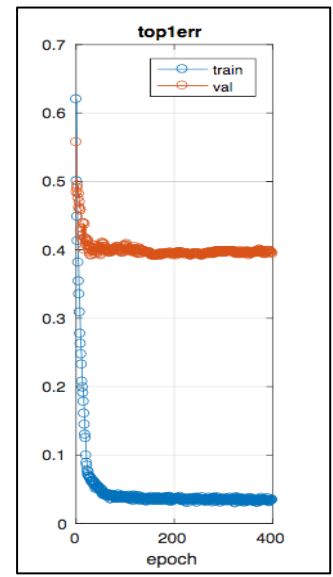

In this experiment, the training top 1 error started out around 0.62 and the validation top 1 error started at 0.55. As the training progressed, accuracy increased. As we see in the graph, the loss stated to decrease in the first 100 iterations. Then it became stable at a value of 0.4 for the validation set and 0.03 for the training set.

Figure 3 shows the confusion matrix of the testing set. The overall testing accuracy was 59%. The classes 'grated', 'juiced' and 'julienne' seem to have the highest precision 73.95%, 68.47% and 73.43% respectively. The class 'Sliced' had the lowest precision of 47.17% and was confused with most of the other classes. Figure 5 shows examples of some misclassified test images of class 'sliced'. One can argue that in the test image to the left of figure 5, the butter cut is rectangular and has the same shape as the definition of diced in the training set (image (a) of figure 4).

Figure 3. Confusion Matrix of the testing set.

| | creamy | diced | grated | juiced | julienne | sliced | whole | Producer Accuracy (Precision) |
|---|---|---|---|---|---|---|---|---|
| creamy | 49 | 4 | 11 | 8 | 0 | 15 | 5 | 53.26% |
| diced | 5 | 53 | 9 | 2 | 4 | 8 | 14 | 55.78% |
| grated | 5 | 10 | 88 | 1 | 7 | 4 | 4 | 73.95% |
| juiced | 11 | 0 | 2 | 63 | 1 | 5 | 10 | 68.47% |
| julienne | 1 | 4 | 4 | 4 | 47 | 2 | 2 | 73.43% |
| sliced | 16 | 21 | 17 | 10 | 14 | 100 | 34 | 47.17% |
| whole | 11 | 9 | 11 | 9 | 6 | 33 | 108 | 57.75% |
| User Accuracy (Recall) | 50% | 52.47% | 61.97% | 64.94% | 59.49% | 59.88% | 61.01% | |

Therefore it is understandable to be classified as diced. Also, the test image to the right of figure 5 looks exactly the same as the training image (b) of figure 4 labeled as 'whole', which can justify why it was mislabeled as 'whole'.

Figure 4. Training images. (a) class 'diced'. (b) class 'whole'

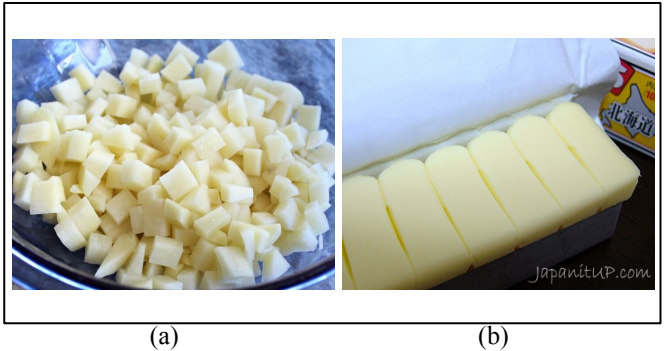

(a)            (b)

Figure 5. Test images from 'sliced' class. (a) misclassified as 'diced'. (b) misclassified as 'whole'

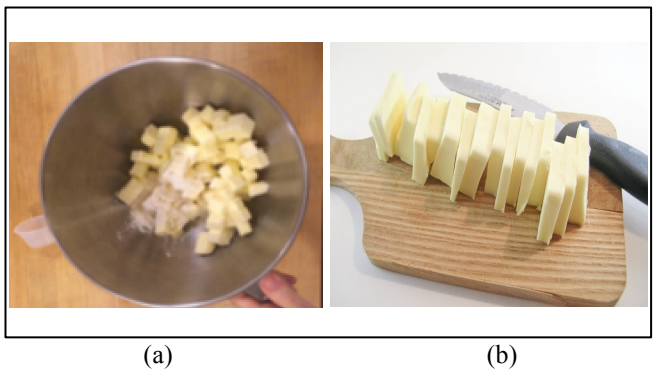

(a)            (b)

Likewise, the class 'creamy' was the second most misclassified class, it was most of the time confused with 'grated' and 'sliced'. Figure 6 shows the training image for the class 'grated' and figure 7 shows the testing images of class 'creamy' misclassified as 'grated'. It can be seen that the image to left of figure 6 is very similar to the image to the left of figure 7 and the images to the right are very similar as well. One can argue that the texture of the image to the right of figure 7 is more grated than creamy. The above-mentioned similarities justify the misclassification.

Figure 6. Training images for grated class.

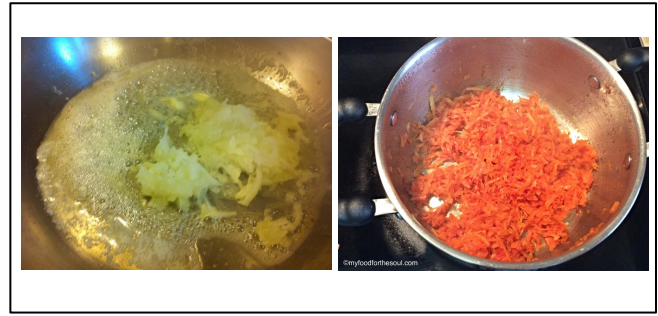

Figure 7. Test images from creamy class misclassified as grated

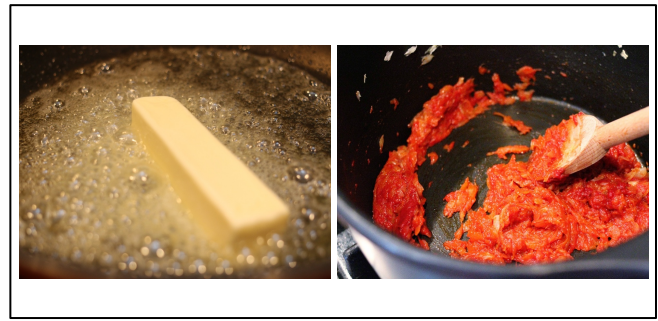

The examples above prove that manual annotation of images with humans in loop is a difficult task. Achieving high annotation precision is challenging.

## VI. CONCLUSION

In this paper, we investigated how to apply deep learning methods to the food state recognition task. The proposed method involves CNNs and transfer learning via fine-tuning. We demonstrated the effectiveness of pre-training and fine-tuning deep convolutional neural network with a small-scale food dataset for state recognition. In future work we plan to explore new ways of reducing the number of weights of deep pre-trained architectures to decrease the effect of over-fitting small datasets. The second aspect that we intend to concentrate on is improving the quality of image annotations using a confidence system based on deep learning.

REFERENCES

[1] K. Yanai and Y. Kawano, "Food image recognition using deep convolutional network with pre-training and fine-tuning," in 2015 IEEE International Conference on Multimedia Expo Workshops (ICMEW), 2015, pp. 1–6.

[2] M. Bosch, F. Zhu, N. Khanna, C. J. Boushey, and E. J. Delp, "combining global and local features for food identification in dietary assessment," IEEE Trans. Image Process. Publ. IEEE Signal Process. Soc., vol. 2011, pp. 1789–1792, 2011.

[3] M.-Y. Chen et al., "Automatic Chinese Food Identification and Quantity Estimation," in *SIGGRAPH Asia 2012 Technical Briefs*, New York, NY, USA, 2012, pp. 29:1–29:4.

[4] Y. Matsuda, H. Hoashi, and K. Yanai, "Recognition of Multiple-Food Images by Detecting Candidate Regions," in *2012 IEEE International Conference on Multimedia and Expo*, 2012, pp. 25–30.

[5] I. Sa, Z. Ge, F. Dayoub, B. Upcroft, T. Perez, and C. McCool, "DeepFruits: A Fruit Detection System Using Deep Neural Networks," *Sensors*, vol. 16, no. 8, Aug. 2016.

[6] K. Fukushima, "Neocognitron: A self-organizing neural network model for a mechanism of pattern recognition unaffected by shift in position," *Biol. Cybern.*, vol. 36, no. 4, pp. 193–202, Apr. 1980.

[7] O. Russakovsky *et al.*, "ImageNet Large Scale Visual Recognition Challenge," *ArXiv14090575 Cs*, Sep. 2014.

[8] Sun, Yu, Yun Lin, and Yongqiang Huang. "Robotic grasping for instrument manipulations." Ubiquitous Robots and Ambient Intelligence (URAI), 2016 13th International Conference on. IEEE, 2016.

[9] Paulius, David, Yongqiang Huang, Roger Milton, William D. Buchanan, Jeanine Sam, and Yu Sun. "Functional object-oriented network for manipulation learning." In Intelligent Robots and Systems (IROS), 2016 IEEE/RSJ International Conference on, pp. 2655-2662. IEEE, 2016.

[10] C. Liu *et al.*, "A New Deep Learning-based Food Recognition System for Dietary Assessment on An Edge Computing Service Infrastructure," *IEEE Trans. Serv. Comput.*, vol. PP, no. 99, pp. 1–1, 2017.

[11] C. Szegedy *et al.*, "Going Deeper with Convolutions," *ArXiv14094842 Cs*, Sep. 2014.

[12] L. Hou, Q. Wu, Q. Sun, H. Yang, and P. Li, "Fruit recognition based on convolution neural network," in *2016 12th International Conference on Natural Computation, Fuzzy Systems and Knowledge Discovery (ICNC-FSKD)*, 2016, pp. 18–22.

[13] L. Pan, S. Pouyanfar, H. Chen, J. Qin, and S. C. Chen, "DeepFood: Automatic Multi-Class Classification of Food Ingredients Using Deep Learning," in *2017 IEEE 3rd International Conference on Collaboration and Internet Computing (CIC)*, 2017, pp. 181–189.

[14] M. Rohrbach, S. Amin, M. Andriluka, and B. Schiele, "A database for fine grained activity detection of cooking activities," in *2012 IEEE Conference on Computer Vision and Pattern Recognition*, 2012, pp. 1194–1201.

[15] K. Chatfield, K. Simonyan, A. Vedaldi, and A. Zisserman, "Return of the devil in the details: Delving deep into convolutional nets," *ArXiv Prepr. ArXiv14053531*, 2014.

[16] A. Vedaldi and K. Lenc, "MatConvNet: Convolutional neural networks for matlab," in *Proceedings of the 23rd Annual ACM Conference on Multimedia Conference*, 2015, pp. 689–692.